\documentclass[lettersize,journal]{IEEEtran}
\usepackage{amsmath,amsfonts}
\usepackage{algorithmic}
\usepackage{array}
\usepackage[caption=false,font=normalsize,labelfont=sf,textfont=sf]{subfig}
\usepackage{textcomp}
\usepackage{stfloats}
\usepackage{url}
\usepackage{verbatim}
\usepackage{graphicx}
\usepackage{color}
\usepackage{soul}
\usepackage{xcolor}
\usepackage{colortbl}
\usepackage{booktabs}
\usepackage{makecell}
\usepackage{newfloat}
\usepackage{listings}
\usepackage{wrapfig}
\usepackage{multirow}
\usepackage{pifont}
\usepackage{caption}
\usepackage{cite}
\def\BibTeX{{\rm B\kern-.05em{\sc i\kern-.025em b}\kern-.08em
    T\kern-.1667em\lower.7ex\hbox{E}\kern-.125emX}}
\usepackage{balance}

\newcommand{\figref}[1]{Fig.~\ref{#1}}
\newcommand{\reqref}[1]{Eq.~\eqref{#1}}
\newcommand{\secref}[1]{Sec.~\ref{#1}}
\newcommand{\tableref}[1]{Table~\ref{#1}}
\definecolor{americanrose}{rgb}{1.0, 0.01, 0.24}

\newcommand{\first}{\textcolor[rgb]{0.91,0.34,0.08}}
\newcommand{\second}{\textcolor[rgb]{0.70,0.08,0.90}}
\usepackage{xspace}
\makeatletter
\DeclareRobustCommand\onedot{\futurelet\@let@token\@onedot}
\def\@onedot{\ifx\@let@token.\else.\null\fi\xspace}
\def\eg{\emph{e.g}\onedot} 
\def\ie{\emph{i.e}\onedot} 
 
\def\etc{\emph{etc}\onedot} 
 
\def\etal{\emph{et al}\onedot}
\makeatother

\definecolor{tabgray}{rgb}{0.83,0.83,0.83}

\usepackage{bibentry}
\begin{document}
\title{IDET: Iterative Difference-Enhanced Transformers for High-Quality Change Detection}
\author{Qing Guo, Ruofei Wang, Rui Huang, Renjie Wan, Shuifa Sun, Yuxiang Zhang
\thanks{
Qing Guo and Ruofei Wang contribute equally to this work and are co-first authors. Rui Huang and Shuifa Sun are the corresponding authors. Q. Guo is with the Center for Frontier AI Research (CFAR) and Institute of High Performance Computing (IHPC), Agency for Science, Technology and Research, Singapore (A*STAR), Singapore (e-mail: tsingqguo@ieee.org). 
R. Wang and R. Wan are with the Department of Computer Science, Hong Kong Baptist University (e-mail:\{ruofei@life., renjiewan@\}hkbu.edu.hk).
R. Huang and Y. Zhang are with the School of Computer Science and Technology, Civil Aviation University of China (e-mail: \{rhuang, yxzhang\}@cauc.edu.cn).
Shuifa Sun is with the School of Information Science and Technology, Hangzhou Normal University, Hangzhou 311121, China (e-mail: watersun@hznu.edu.cn). Copyright reserved by IEEE TETCI.
}}

\markboth{Journal of \LaTeX\ Class Files,~Vol.~14, No.~8, August~2021}%
{Shell \MakeLowercase{\textit{et al.}}: A Sample Article Using IEEEtran.cls for IEEE Journals}


\maketitle

\begin{abstract}
Change detection~(CD) is a crucial task in various real-world applications, aiming to identify regions of change between two images captured at different times. However, existing approaches mainly focus on designing advanced network architectures that map feature differences to change maps, overlooking the impact of feature difference quality. In this paper, we approach CD from a different perspective by exploring \textit{how to optimize feature differences to effectively highlight changes and suppress background regions}. To achieve this, we propose a novel module called the iterative difference-enhanced transformers (IDET).
IDET consists of three transformers: two for extracting long-range information from the bi-temporal images, and one for enhancing the feature difference. Unlike previous transformers, the third transformer utilizes the outputs of the first two transformers to guide iterative and dynamic enhancement of the feature difference. To further enhance refinement, we introduce the multi-scale IDET-based change detection approach, which utilizes multi-scale representations of the images to refine the feature difference at multiple scales. Additionally, we propose a coarse-to-fine fusion strategy to combine all refinements.
Our final CD method surpasses nine state-of-the-art methods on six large-scale datasets across different application scenarios. This highlights the significance of feature difference enhancement and demonstrates the effectiveness of IDET. Furthermore, we demonstrate that our IDET can be seamlessly integrated into other existing CD methods, resulting in a substantial improvement in detection accuracy. Code is available at \url{https://github.com/rfww/IDET.}
\end{abstract}

\section{Introduction}
\label{sec:intro}

\begin{figure}[!h]
    \centering
    \includegraphics[width=1.0\linewidth]{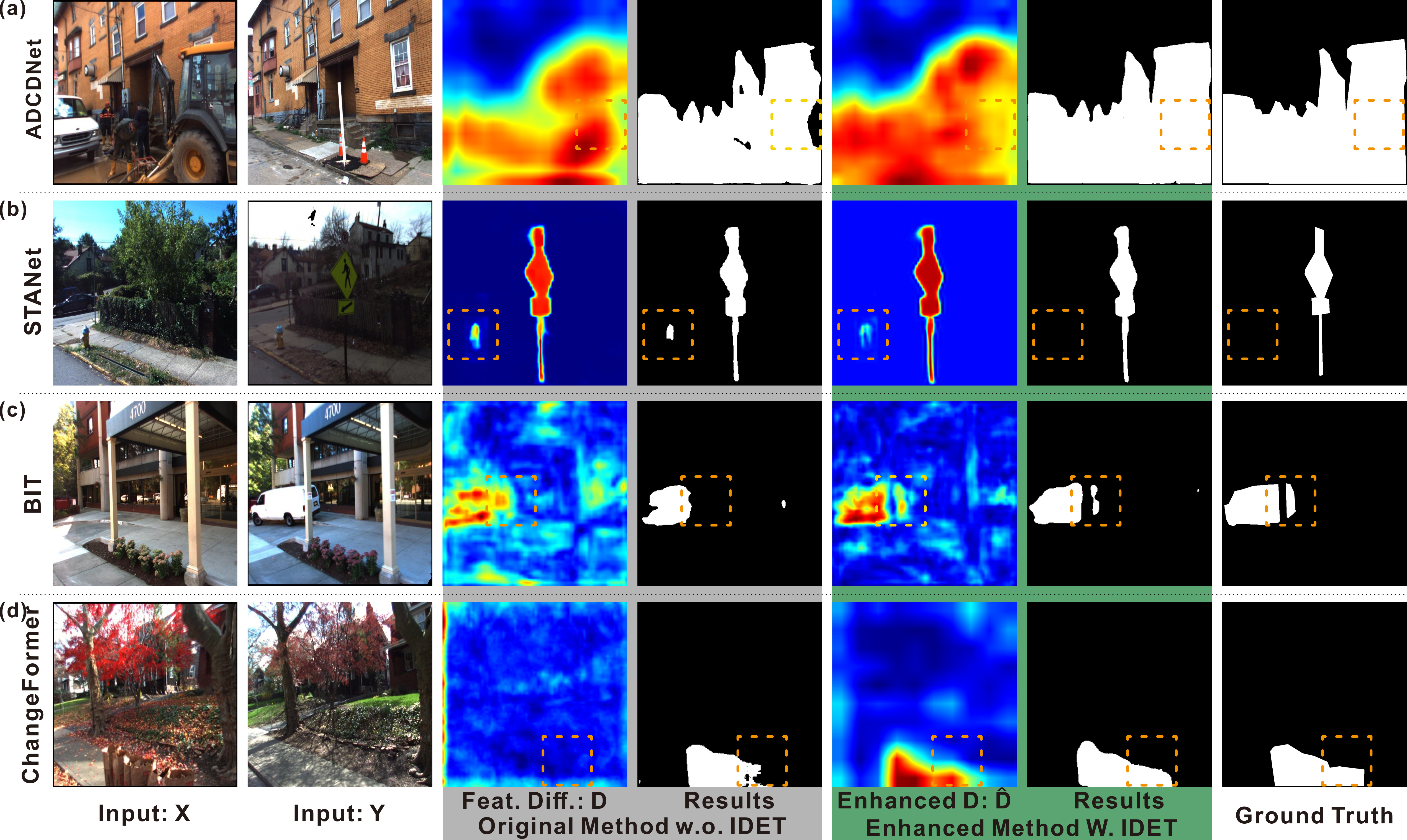}
    \caption{
    Visualization comparison of four existing change detectors (\ie, ADCDNet \cite{huang2020change}, STANet \cite{chen2020spatial}, BIT \cite{chen2021a} and ChangeFormer \cite{bandara2022transformer}) and their enhanced counterparts with our IDET. The first two columns are the inputs. The third and fourth columns are the feature differences (\ie, $\textbf{D}$) and detection results of the four existing methods. The fifth and sixth columns are enhanced feature differences (\ie, $\hat{\textbf{D}}$) and the corresponding detection results. The final column displays the ground truth.}
    
    \label{fig:motivation}
\end{figure}

\IEEEPARstart{C}{hange} detection (CD) aims to identify areas of change resulting from object variations, such as the appearance of new objects in a scene, in images captured at different times. This technique finds wide applications in urban development \cite{buch2011review,barkur2022rscdnet}, disaster assessment \cite{brunner2010change,zhang2022unsupervised}, resource monitoring and utilization \cite{khan2017forest,mou2018learning}, abandoned object detection \cite{lin2015abandoned}, illegally parked vehicles \cite{albiol2011detection} and military \cite{liu2013intelligent,Liang2017adaptive}. Change detection is a challenging task due to variations in angles, illuminations, and unknown changes in the background between the two images. For instance, in \figref{fig:motivation}~(b), the sky color and tree intensity differ significantly in the two images, even though they should not be identified as changes.

\begin{figure*}[!h]  
\centering  
\includegraphics[width=1.0\linewidth]{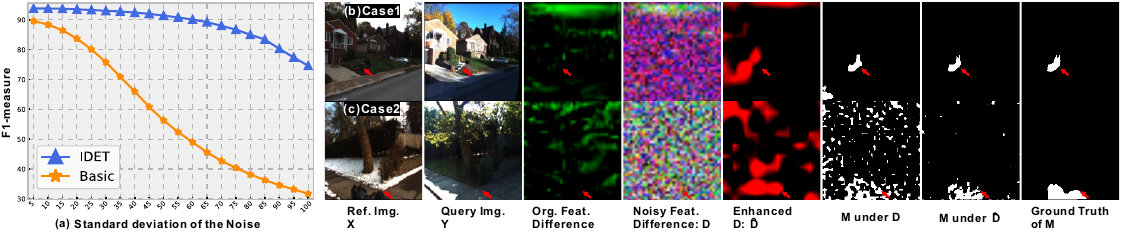} 
\caption{(a) Evaluation results on the VL-CMU-CD dataset by adding noise with different severities to the feature difference $\mathbf{D}$. (b) and (c) display two cases including the original feature difference, noisy feature difference $\mathbf{D}$, enhanced feature difference $\hat{\mathbf{D}}$, the detection results $\mathbf{M}$ under $\mathbf{D}$ and $\hat{\mathbf{D}}$, and the ground truth, respectively. The main changes are highlighted via red arrows. IDET effectively suppresses the background regions and highlights changes.}  
\label{fig:analysis}  
\end{figure*}

A straightforward solution for detecting changes is to subtract the features of the two images and map the absolute feature difference to the final change mask \cite{zhang2020feature,wang2022high,chen2020spatial,chen2021a}. These methods aim to extract more distinct features to overcome background disturbance and position misalignment. For example, Chen \etal \cite{chen2020spatial} propose a Siamese-based spatial-temporal attention neural network (STANet), which utilizes self-attention to model spatial-temporal information in the input images. They also propose a transformer-based approach, namely bitemporal image transformer (BIT) \cite{chen2021a}, which effectively models spatial-temporal contexts. Additionally, Bandara \etal \cite{bandara2022transformer} implement a transformer-based change detector called ChangeFormer, with a hierarchically structured transformer-based encoder and a multi-layer perception for the decoder. Although these methods achieve impressive results, they still produce erroneous detection results when the features of the input images are not discriminative across change and unchanged regions. As shown in \figref{fig:motivation}, the above state-of-the-art (SOTA) methods may miss change regions or produce false changes due to low-quality feature differences. For instance, STANet incorrectly labels the fire hydrant as a changed object, and ADCDNet \cite{huang2020change} misses some changing pixels. Even the SOTA transformer-based methods, BIT and ChangeFormer, face similar issues. These sub-optimal detection performances show that the noisy background regions between bi-temporal images can easily mislead the CD detector to detect changes.

To address the aforementioned limitations, we propose a novel method to improve the quality of the feature difference, thereby highlighting change regions and suppressing confused backgrounds. We introduce a new module called iterative difference-enhanced transformers (IDET), which not only achieves higher detection accuracy within a self-contained framework but also enhances existing methods (refer to the fifth and sixth columns in \figref{fig:motivation}). IDET comprises three transformers: the first two extract global information from the input images, while the third enhances the feature differences. Notably, the third transformer differs from existing transformers as it iteratively and dynamically refines the feature difference using the outputs of the first two transformers as guidance. Additionally, we propose a multi-scale IDET-based detection method that employs multi-scale representations for IDET to enable multi-scale refinements. Extensive experiments conducted on six CD datasets demonstrate the effectiveness of the proposed method, showing the great significance of the feature difference quality. 

Overall, our contributions can be summarized as follows:
\begin{itemize}
   
 \item We develop a basic feature difference-based CD method and investigate the influence of feature difference quality on detection accuracy by introducing random noise with varying severities. Our analysis demonstrates the critical role of feature difference quality in achieving high-quality change detection.
\item We propose an iterative difference-enhanced transformer (IDET) that refines the feature difference by extracting long-range information from the inputs as guidance. Furthermore, we extend IDET to a multi-scale approach that incorporates multi-scale representations and a coarse-to-fine fusion strategy.
\item We conduct extensive experiments on six datasets, encompassing diverse scenarios, resolutions, and change ratios. IDET outperforms nine state-of-the-art methods. 
\item We integrate IDET into four state-of-the-art methods (ADCDNet \cite{huang2020change}, STANet \cite{chen2020spatial}, BIT \cite{chen2021a} and ChangeFormer \cite{bandara2022transformer}) and demonstrate significant improvements in their detection accuracy. These results highlight the generalization and flexibility of our method.

\end{itemize}

\section{Related Work}
\subsection{Change detection methods}
Change detection (CD) methods can be categorized into two groups based on their applications: CD for common scenarios and remote sensing fields. Both types detect changes from a pair of images, but there are differences in terms of image resolution and the size of the change regions. CD methods for common scenarios \cite{guo2018learning} typically use images captured by consumer cameras, which may result in camera pose misalignment and illumination differences \cite{atghaei2022industrial}. In contrast, remote sensing CD methods rely on high-resolution remote sensing images \cite{chen2020dasnet,zhang2021escnet}, hyper-spectral images \cite{2018GETNET}, SAR images \cite{wu2021commonality}, heterogeneous images \cite{sun2021structure}, \etc. The changes between bi-temporal images tend to be quite nuanced and understated, making them difficult to detect. Early studies have demonstrated the effectiveness of principal component analysis (PCA) \cite{li1998principal}, change vector analysis (CVA) \cite{malila1980change}, and k-means~\cite{gupta2019change} for CD. However, with the advancement of deep neural networks, recent methods have also incorporated them as the mainstream solution to conduct CD.

Various methods have been proposed for change detection (CD) using machine learning techniques, including data augmentation~\cite{chen2021adversarial,huang2023background}, pseudo labels generation~\cite{peng2020semicdnet}, transfer learning~\cite{chen2022semantic}, \etc.
Dual correlation attention modules are utilized to refine the change features at the same scale and then fuse together to generate a change map \cite{zhang2021object}.
Another notable method is the triplet network proposed by Nguyen \etal \cite{nguyen2018change}, which leverages motion features extracted from small image patches to predict CD. FgSegNet \cite{gao2021extracting} is a post-processing algorithm that optimizes coarse change detection results by incorporating the edges of changed objects. Additionally, MCRCNN \cite{santos2020scene} introduces a multiscale cascade residual convolutional neural network composed of a background network, a residual processing module, and a segmentation network, to predict a binary change map.

To address heterogeneity in CD images, Luppino \etal \cite{luppino2022code} employ two auto-encoders to align bi-temporal images in code space and then detect changes from two this space via existing techniques. These works primarily concentrate on designing novel model architectures to enhance detection performance. Attaching more attention to effective feature extraction~\cite{sun2024spatial} and feature fusion~\cite{gao2024relating}.

In contrast, our focus lies on a new view: the feature difference quality. We study how the feature difference quality impacts the final change map prediction. Specifically, the high-quality feature differences can effectively suppress background regions and highlight changes, while low-quality counterparts would result in false detection. Through extensive research, we propose a feature difference enhancement method that outperforms SOTA methods in various applications, ranging from common scenarios to remote sensing environments.
%
%
\begin{figure*}[t]
    \centering
    \includegraphics[width=0.75\linewidth]{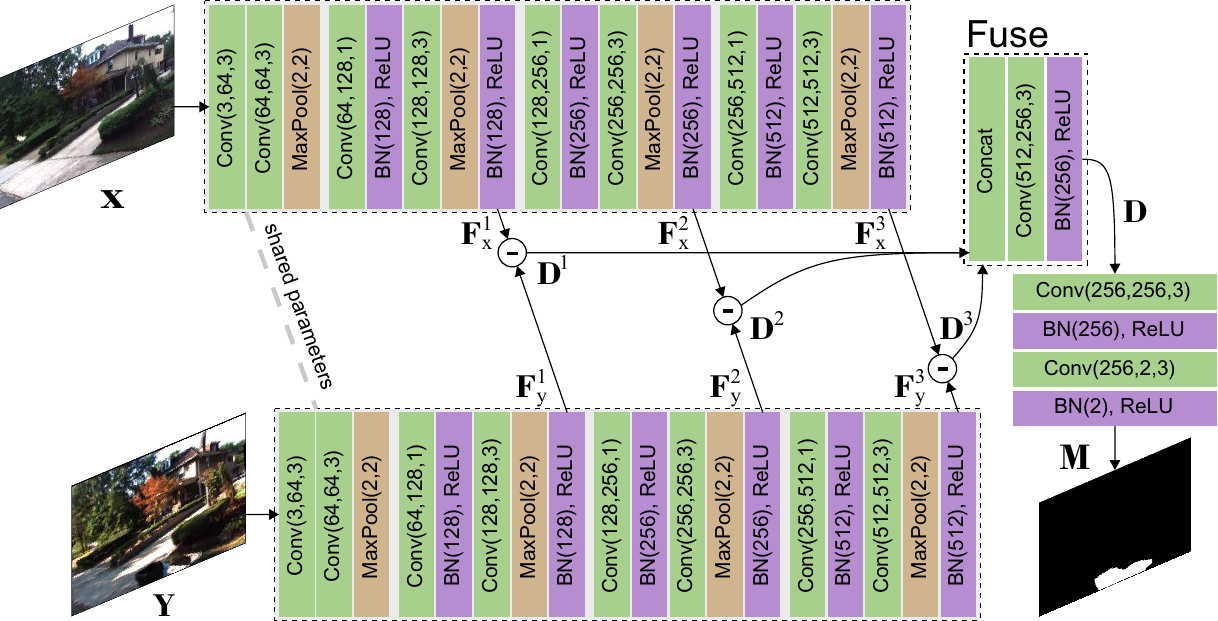}
    \caption{ Existing architecture for feature difference-based change detection.}
    \vspace{-5px}
    \label{fig:basic}
\end{figure*}
%

\subsection{Transformer for vision tasks}
The transformer has gained popularity in natural language processing (NLP) tasks due to its capability to handle long-range dependencies, as initially proposed in~\cite{vaswani2017attention}. Recently, transformers have shown promising results in computer vision applications, including image classification~\cite{dosovitskiy2020image,sun2021boosting}, object detection~\cite{carion2020end,touvron2021training,zhu2020deformable}, image captioning~\cite{cornia2020meshed}, semantic segmentation~\cite{zheng2021rethinking,liu2021swin,dong2021cswin,xie2021segformer}, and \etc. 
Given their impressive performance, it is natural to explore the potential of transformers in change detection (CD). 

BIT \cite{chen2021a} employs the transformer to learn a set of special semantic tokens from bi-temporal images to conduct CD, which assumes that the changes between bi-temporal images can be represented by a few visual words. Nonetheless, learning semantic tokens from such small and varied objects presents a significant challenge.
ChangeFormer \cite{bandara2022transformer} hierarchically stacks four transformer encoders to extract long-range features for generating change masks, whereas the small resolution of encoders often fails to capture detailed boundary changes.
Sun~\cite{sun2024spatial} propose employing transformers to refine the spatiotemporal relationships within the same-scale bi-temporal features to extract image representations.
All these methods focus on designing the novel architecture to optimize feature extraction or fusion to improve detection performance, while overlooking the impact caused by the change-related feature, \ie, feature difference.

To this end, we approach CD from the view of the feature difference quality and propose the IDET to effectively enhance feature difference. 
Specifically, our proposed IDET consists of three transformers: the first two optimize bi-temporal image features, while the last one is designed to refine feature differences. 
Moreover, IDET can be viewed as a plug-and-play module that can be incorporated into other off-the-shelf methods to enhance their performance.

\section{Feature Difference-based CD and Motivation}
\label{sec:featuremot}

In this section, we first introduce a lightweight convolutional neural network, the basic feature difference-based CD model, and then analyze the issues from the viewpoint of the feature difference quality.

\subsection{Feature Difference-based CD}
\label{subsec:featdiff}

Given a reference image $\mathbf{X}$ and a query image $\mathbf{Y}$, a change detector $\phi(\cdot)$ takes the two images as the input and predicts a change map $\mathbf{M}$ that indicates changing regions between the two images, which are caused by object variations (\eg, new objects appearing in the scene) instead of the environment changes (\eg, light variations).
We can represent the whole process as $\mathbf{M}=\phi(\mathbf{X},\mathbf{Y})$.
The main challenge is how to distinguish the object changes from environmental variations.
Previous works calculate the difference between the features of $\mathbf{X}$ and $\mathbf{Y}$ explicitly and map it to the final change mask, which is robust to diverse environment variations~\cite{huang2020change}. 
We follow the feature difference solution and build the \textit{basic change detection method} via a CNN.
Specifically, we feed $\mathbf{X}$ or $\mathbf{Y}$ to the CNN and get a series of features $\{\mathbf{F}^{l}_{\text{x}~\text{or}~\text{y}}\}_{l=1}^L$.
Then, we can calculate the feature difference at each layer $l$ and fuse them by a convolution layer
%
\begin{align}\label{eq:featdiff}
\mathbf{D}= \text{Fuse}(\{\mathbf{D}^{l}\}_{l=1}^L),~\forall~l~, \mathbf{D}^{l}=|\mathbf{F}^{l}_{\text{x}}-\mathbf{F}^{l}_{\text{y}}|, 
\end{align}
%
where the feature difference $\mathbf{D}$ can be fed to the other two convolution layers to predict the final change map $\mathbf{M}$. We present all architecture details in \figref{fig:basic}. Previous works mainly design more advanced architecture to replace the basic CNN \cite{zhang2020deeply}. In contrast, we focus on the influence of the quality of $\mathbf{D}$, which is not explored yet.

\subsection{Analysis and Motivation}
\label{subsec:analysis}

Based on \figref{fig:basic}, we study the importance of the quality of $\mathbf{D}$ to the final change map. 
Given an image pair from VL-CMU-CD dataset ~\cite{alcantarilla2018street} (\eg, \figref{fig:analysis}~(b) and (c)), we feed them to \reqref{eq:featdiff} and get the original feature difference denoted as $\mathbf{D}_\text{o}$. Then, we add random noise to the unchanged regions of $\mathbf{D}_\text{o}$ to simulate the feature difference with diverse qualities. Specifically, according to the ground truth of the change regions, we first calculate the averages of feature difference values inside and outside change regions, which are denoted as $d_\text{o}^\text{inside}$ and $d_\text{o}^\text{outside}$, respectively. Then, we set a zero-mean random noise with standard deviation as $\alpha|d_\text{o}^\text{inside}-d_\text{o}^\text{outside}|$ and add the noise to the unchanged regions in $\mathbf{D}_\text{o}$ to get a degraded version denoted as $\mathbf{D}$. 
We can conduct this process for all examples with the $\alpha$ from 5 to 100 with the interval of 5.
Then, for each $\alpha$, we can evaluate the detection results via the F1-measure.
According to the dark orange stars in \figref{fig:analysis}~(a), we see that the basic CD method is very sensitive to the quality of the feature difference. Clearly, as the noise becomes larger, the F1 value decreases significantly. We have a similar conclusion on the visualization results in \figref{fig:analysis}: the low-quality feature difference leads to incorrect detection.

\begin{figure}[t]
    \centering
    \includegraphics[width=0.97\linewidth]{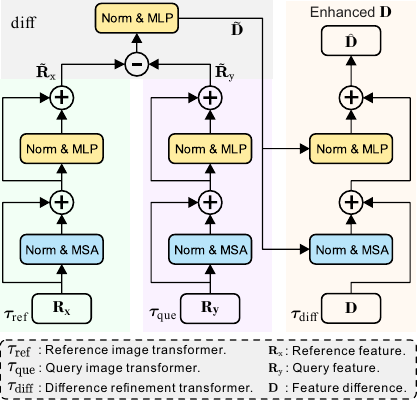} 
    \caption{The architecture of the Iterative Difference-enhanced Transformers (IDET) with iteration $T=1$. With inputting the reference and query image representations~(\ie, $\mathbf{R}_{\text{x}}$ and $\mathbf{R}_{\text{y}}$), the initial feature difference $\mathbf{D}$ would be refined by the $\Tilde{\mathbf{D}}$ to generate the enhanced counterpart $\hat{\mathbf{D}}$, where $\mathbf{D}$ and $\Tilde{\mathbf{D}}$ are obtained by \reqref{eq:featdiff} and \reqref{eq:idet_diff}, respectively. In the next iteration~(\eg, $T=2$), $\mathbf{R}_{\text{x}}$, $\mathbf{R}_{\text{y}}$, and $\mathbf{D}$ would be replaced by $\Tilde{\mathbf{R}}_{\text{x}}$, $\Tilde{\mathbf{R}}_{\text{y}}$, and $\hat{\mathbf{D}}$ respectively, enhancing $\hat{\mathbf{D}}$ again.}
    \label{fig:idet}
\end{figure}
\section{Methodology}
\label{sec:method}

\subsection{Problem Definition and Challenges}
\label{subsec:problem}

Due to the importance of feature difference quality, we formulate the feature difference enhancement as a novel task: given the representations of $\mathbf{X}$ and $\mathbf{Y}$ (\ie, $\mathbf{R}_{\text{x}}$ and $\mathbf{R}_{\text{y}}$) and an initial feature difference $\mathbf{D}$, we aim to refine $\mathbf{D}$ and get a better counterpart (\ie, $\hat{\mathbf{D}}$) where the object changes are highlighted and other distractions are suppressed. 

The feature difference enhancement is a non-trivial task since pixel changes between $\mathbf{X}$ and $\mathbf{Y}$ stem from the object and environment changes, making it difficult to disentangle and distinguish the changes from each other. 
To achieve this enhancement, a feature difference enhancement method should first understand the whole scene in both reference and query images, identifying which regions are caused by object changes. Moreover, the extracted discriminative information should be effectively embedded into the initial feature difference $\mathbf{D}$ to guide the enhancement. 

\begin{figure}[t]

    \includegraphics[width=\linewidth]{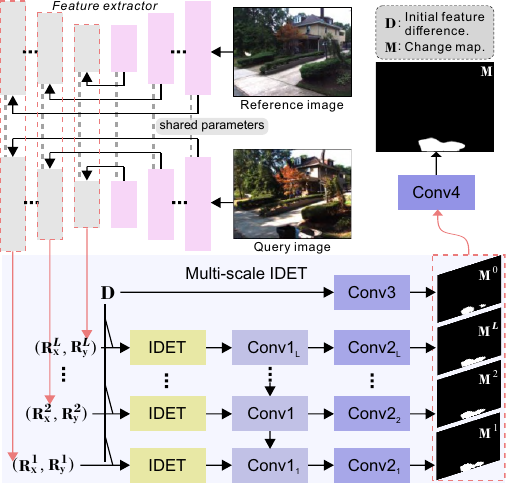} 
    \caption{The framework of our multi-scale IDET-based change detection. First, we input bi-temporal images~(\ie, reference image and query image) into a feature extractor to extract multi-scale features ($\mathbf{R}_{\text{x}}^l$, $\mathbf{R}_{\text{y}}^l$).
    Second, IDET is used to enhance feature difference $\mathbf{D}$ at different scales. Finally, a coarse-to-fine fusion strategy is introduced to fuse all enhanced differences to generate the final change map $\mathbf{M}$.}
    \label{fig:msidet}
\end{figure}

\subsection{Iterative Difference-enhanced Transformers}
\label{subsec:idet}

\textbf{Overview.} We employ the transformer to achieve the feature difference enhancement due to the impressive improvements of transformer-based methods in diverse vision tasks \cite{xie2021segformer}. Note that, the transformer is able to
capture the long-term dependency of the whole inputs, which meets the requirements of the feature difference enhancement. 
Nevertheless, how to adapt such an advanced technique to the new task is not clear. To this end, we propose the \textit{iterative difference-enhanced transformers (IDET)} for feature difference enhancement.

IDET contains three transformers, \ie, reference image transformer $\tau_{\text{ref}}(\cdot)$, query image transformer $\tau_{\text{que}}(\cdot)$, and difference enhancement transformer $\tau_{\text{diff}}(\cdot)$. The whole process of IDET is shown in \figref{fig:idet} and can be formulated as:
%
\begin{align}\label{eq:idet}
\hat{\mathbf{D}}=\tau_{\text{diff}}(\mathbf{D},\Tilde{\mathbf{D}}),~\text{with}~\Tilde{\mathbf{D}} =\text{diff}(\Tilde{\mathbf{R}}_\text{x},\Tilde{\mathbf{R}}_\text{y}),
\end{align}
%
where $\Tilde{\mathbf{R}}_\text{x}$ and $\Tilde{\mathbf{R}}_\text{y}$ are the enhanced versions of image representations $\mathbf{R}_\text{x}$ and $\mathbf{R}_\text{y}$ refined by transformers $\tau_{\text{ref}}(\cdot)$ and $\tau_{\text{que}}(\cdot)$, respectively~(\ie, $\Tilde{\mathbf{R}}_{\text{x}}=\tau_{\text{ref}}(\mathbf{R}_\text{x})$, $\Tilde{\mathbf{R}}_{\text{y}}=\tau_{\text{que}}(\mathbf{R}_\text{y})$).
The function $\text{diff}(\Tilde{\mathbf{R}}_\text{x},\Tilde{\mathbf{R}}_\text{y})$ is defined as:
%
\begin{equation} \label{eq:idet_diff}
        \Tilde{\mathbf{D}}=\text{diff}(\Tilde{\mathbf{R}}_\text{x},\Tilde{\mathbf{R}}_\text{y}) 
        =\text{MLP}(\text{Norm}(|\tau_{\text{ref}}(\mathbf{R}_\text{x})-\tau_{\text{que}}(\mathbf{R}_\text{y})|)),
\end{equation}
which aims to generate difference information (\ie, $\Tilde{\mathbf{D}}$) from the two long-range features (\ie, $\tau_{\text{ref}}(\mathbf{R}_\text{x})$ and $\tau_{\text{que}}(\mathbf{R}_\text{y})$) via a multilayer perceptron (MLP) and a layer norm (Norm). Finally, the $\Tilde{\mathbf{D}}$ is fed to the difference enhancement transformer  (\ie, $\tau_{\text{diff}}(\cdot)$) to guide the enhancement of the initial difference $\mathbf{D}$.
\textit{Moreover, we can make these enhanced features~(\ie, $\Tilde{\mathbf{R}}_\text{x}$, $\Tilde{\mathbf{R}}_\text{y}$, $\hat{\mathbf{D}}$) as the inputs of next iteration to enhance the feature difference~(\eg, $\hat{\mathbf{D}}$) again. As a result, the method becomes iterative.}
We show the architecture of the proposed IDET in \figref{fig:idet} and detail them in the following subsections.

\textbf{Architectures of $\tau_{\text{ref}}(\cdot)$ and $\tau_{\text{que}}(\cdot)$.} We eliminate the position token and class token in the model designation because we don't use patch embedding and ignore the object labels. 
The whole process can be formulated as
%
\begin{equation}\label{eq:idet_former12}
    \begin{split}
        \tau_{*}(\mathbf{R}_{*}) &= \text{MLP}(\text{Norm}(\mathbf{Z}))+\mathbf{Z},~\text{with}~ \\
        \mathbf{Z} &=\text{MSA}(\text{Norm}(\mathbf{R}_{*}))+\mathbf{R}_{*},
    \end{split}
\end{equation}
%
where $\text{MSA}(\cdot)$ is the efficient multi-head self-attention introduced in \cite{xie2021segformer}.

\textbf{Architecture of $\tau_{\text{diff}}(\cdot)$.} In contrast to the above transformers,
$\tau_{\text{diff}}(\cdot)$ uses the difference information $\Tilde{\mathbf{D}}$ obtained by \reqref{eq:idet_diff} to guide the enhancement of the initial $\mathbf{D}$. To this end, we modify the \reqref{eq:idet_former12} by
%
\begin{equation}\label{eq:idet_former3}
    \begin{split}
        \tau_\text{diff}(\mathbf{D},\Tilde{\mathbf{D}}) &= \text{MLP}(\text{Norm}(\Tilde{\mathbf{D}}))+\mathbf{Z},~\text{with}~ \\
        \mathbf{Z} &=\text{MSA}(\text{Norm}(\Tilde{\mathbf{D}}))+\mathbf{D}.
    \end{split}
\end{equation}
%
The intuitive idea is to use MSA and MLP to extract the long-term information in $\Tilde{\mathbf{D}}$ and embed it to update $\mathbf{D}$.

\subsection{Multi-scale IDET-based CD}

With the proposed IDET, we are able to update the aforementioned basic feature difference-based CD methods. The key challenge lies in how to select the representations (\ie, $\mathbf{R}_\text{x}$ and $\mathbf{R}_\text{y}$ in \reqref{eq:idet}) to refine the feature difference $\mathbf{D}$ in \reqref{eq:featdiff}, and how to fuse refined results from different representations. 
Here, we propose to use UNet~\cite{ronneberger2015u} to extract multi-scale convolutional features as the representations, enabling us to obtain multiple refined feature differences. Subsequently, a coarse-to-fine fusion strategy is proposed to combine multi-scale results for final change detection.
Specifically, we employ a UNet feature extractor to extract  multi-scale representations from $\mathbf{X}$ and $\mathbf{Y}$, respectively. The multi-scale representations are denoted as $\{\mathbf{R}^l_\text{x}\}^{L}$ and $\{\mathbf{R}^l_\text{y}\}^{L}$, where $L$ denotes the number of scales. 
Note that, given $\mathbf{R}^l_\text{x}$ and $\mathbf{R}^l_\text{y}$, we still cannot refine $\mathbf{D}$ through the IDET directly since $\mathbf{D}$ from \reqref{eq:featdiff} has smaller resolution than $\mathbf{R}^l_\text{x}$ and $\mathbf{R}^l_\text{y}$. 
To use IDET at the scale $l$, we first upsample $\mathbf{D}$ to the size of $\mathbf{R}^l_\text{x}$ or $\mathbf{R}^l_\text{y}$ and then refine to get a enhanced counterpart $\hat{\mathbf{D}}^l$. For all scales, we have $\{\hat{\mathbf{D}}^l\}_{l=1}^L$ and fuse them via a coarse-to-fine strategy, \ie,
%
\begin{align}\label{eq:msidet_fusion1}
\hat{\mathbf{D}}'^l = \text{Conv1}_l([\hat{\mathbf{D}}^l,\hat{\mathbf{D}}'^{(l-1)}]),~\text{with}~l=1,\ldots,L,
\end{align}
%
where `$[\cdot]$' denotes the concatenation operation and $~\hat{\mathbf{D}}'^{0}=\text{empty}$. For $\hat{\mathbf{D}}'^l$ and the initial $\mathbf{D}$, we use a convolution layer to extract a change map:
%
\begin{align}\label{eq:msidet_fusion2}
\mathbf{M}^l =\text{Conv2}_l(\hat{\mathbf{D}}'^l),~\text{and}~\mathbf{M}^0 = \text{Conv3}(\mathbf{D}).
\end{align}
%
Then, the final change map can be obtained via $\mathbf{M}=\text{Conv4}([\mathbf{M}^0,\ldots,\mathbf{M}^L]).$

\subsection{Implementation Details}
\label{subsec:impl}

{\bf Architecture of UNet.} As shown in \figref{fig:basic}, we adopt the VGG16~\cite{simonyan2014very} as the encoder, which has been divided into five squeeze modules to extract abstract features by reducing the spatial resolution. For the decoder, we use two convolutional layers with normalization layers and ReLU to 
reconstruct the feature resolution by integrating high-resolution and low-level features from squeeze modules layer by layer.

{\bf Setups of Conv* for fusion.} 
$\text{Conv1}_*(\cdot)$ in \reqref{eq:msidet_fusion1} contains three modules, each of which has a convolutional layer and a ReLU layer. The first convolution is responsible for reducing the channel number to half of the channel number of the input feature. The second convolution has the same input and output channel number to further refine the feature. The third convolution produces a feature with a given output channel number. The filter size in convolutions is $3\times3$. $\text{Conv2}_*(\cdot)$ in \reqref{eq:msidet_fusion2} is a convolutional layer with the size of $N \times 3 \times 3  \times 2$, where $N$ represents the input channel number.
$\text{Conv3}(\cdot)$ is used to generate change map $\mathbf{M}^0$ from the feature difference $\mathbf{D}$. $\text{Conv4}(\cdot)$ consists of $12\times3\times3\times 2$ convolutional filters to generate final change map $\mathbf{M}$.

{\bf Loss functions.} 
Given a prediction $\mathbf{M}^*$, we adopt the cross-entropy loss function to calculate their error distance between $\mathbf{M}^*$ and ground truth, denoting as $\mathcal{L}_{\mathbf{M}^*}$. In addition, we also compute the loss on multi-scale binary maps to ensure convergence, thereby the total loss can be obtained by:
\begin{align}
  \mathcal{L} =\sum_{l=1}^{L}\lambda_l\mathcal{L}_{\mathbf{M}^l}+\lambda_M\mathcal{L}_\mathbf{M},
  \label{eq:loss_fuctions}
\end{align}
where $\mathcal{L}_{\mathbf{M}^l}$ denotes the loss between a single-scale map and ground truth. The parameter $\lambda$ is set to $\lambda_l=\lambda_M=1$.

{\bf Training parameters.} 
Our model is implemented on PyTorch~\cite{paszke2017automatic} and trained with a single NVIDIA Geforce 2080Ti
GPU. The basic learning rate is set to 1e-3. We set the batch size to 4. The parameters
are updated by the Adam algorithm with a momentum of 0.9 and weight decay of 0.999. Note that we use the same training parameters for all six datasets. Following \cite{chen2020spatial}, we train all CD methods with 20 and 200 epochs for common scene datasets and remote sensing datasets, respectively.


\section{Experimental Results}
\label{sec:expermients}

\subsection{Setups}
\textbf{Baselines.} We compare our method with seven CD methods, i.e., FCNCD~\cite{long2015fully}, ADCDNet~\cite{huang2020change}, CSCDNet~\cite{sakurada2020weakly}, BIT~\cite{chen2021a}, IFN~\cite{zhang2020deeply}, STANet~\cite{chen2020spatial}, ChangeFormer \cite{bandara2022transformer}, ARPPNet~\cite{chen2018learning}, STMSNet~\cite{zhou2022spatial}, SFBNet~\cite{sun2024spatial}, and RCTNet~\cite{gao2024relating}.
All compared methods are trained under the same setup. 

\begin{figure*}[t]  
\centering  
\includegraphics[width=1.0\linewidth]{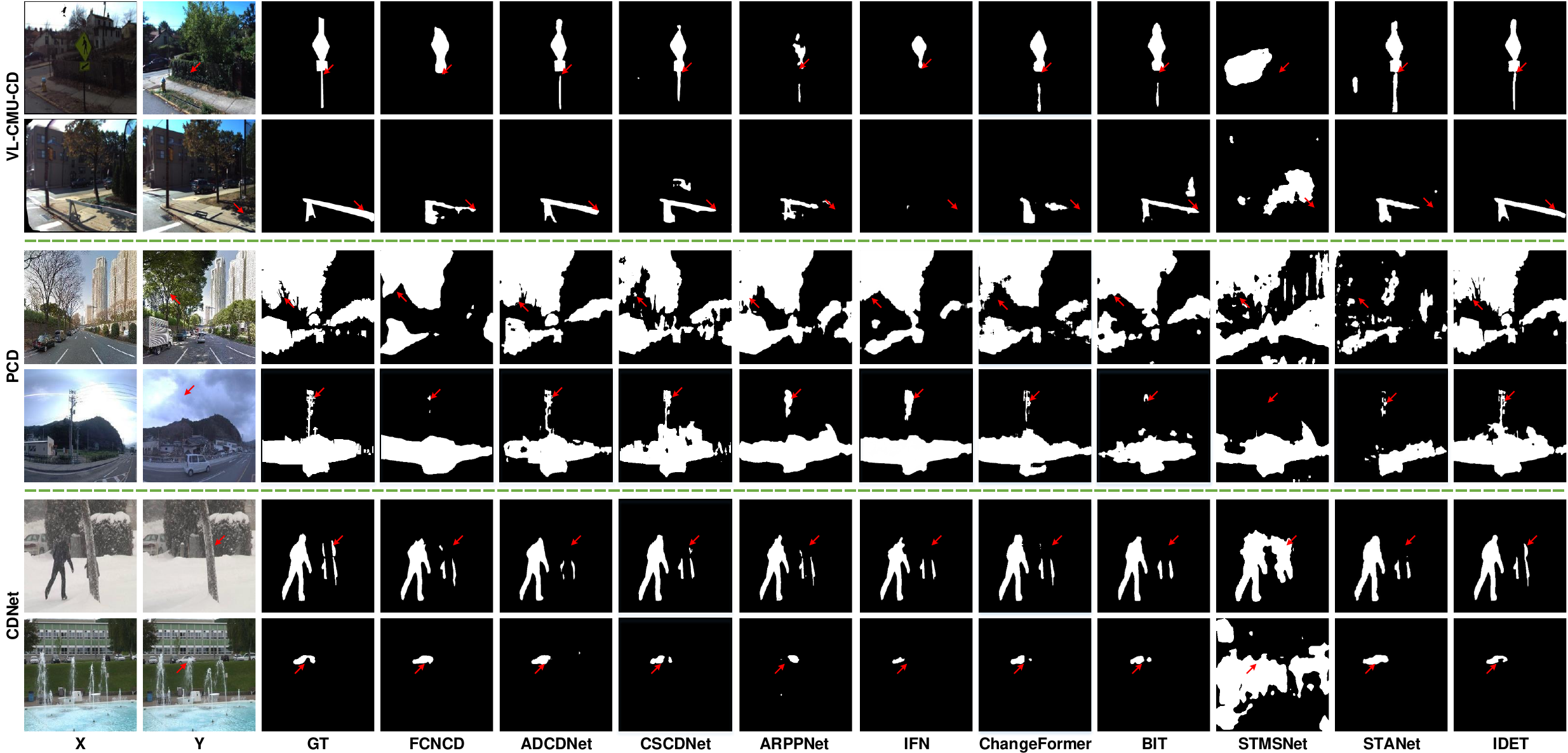}  
\caption{Detection results of different CD methods on VL-CMU-CD~\cite{alcantarilla2018street}, PCD~\cite{sakurada2015change} and CDNet~\cite{goyette2012changedetection} datasets.
}  
\label{fig:commons}  
\end{figure*}

\begin{table*}[t]
    \centering
    \caption{Dataset details. The change ratio is the proportion of the number of changed pixels to the total number of pixels.}
    \resizebox{0.9\linewidth}{!}{
    \begin{tabular}{l|c|c|c|c|c|c}
    \toprule
    & VL-CMU-CD & PCD & CDNet & LEVIR-CD & CDD & AICD \\
    \midrule
    Image number & 1362 & 200 & 50209 & 637 & 11 & 1000 \\
    \multirow{2}{*}{Resolution}& \multirow{2}{*}{$1024\times 768$} &
    \multirow{2}{*}{$1024\times 224$} & 
    $320\times 240$, &
    \multirow{2}{*}{$1024\times 1024$} &
    $4725\times 2700$ &
    \multirow{2}{*}{$800\times 600$} \\
    &&& $720\times 576$ && $1900\times 1000$ & \\

    \makecell[l]{Categories} &
    \makecell[c]{bins, signs, \\ maintenance, \\ vehicles, \etc}&
    \makecell[c]{trees, \\ buildings, \\ vehicles, \etc} &
    \makecell[c]{person,\\ vehicles, \\boats, \etc} &
    buildings&
    \makecell[c]{buildings, \\vehicles, \\roads, \etc} & \makecell[c]{pavilions, \\ships \etc}\\
    
    Change ratio &0.0716&0.2587&0.0479&0.1116&0.1234&0.0022 \\
    Training image number &16005&8000&40148&18015&12998&25074 \\
    Testing image number &332&120&10061&4674&3000&464 \\
    Training image size &$320 \times 320$&$320 \times 320$&$320 \times 320$&$256 \times 256$&$256 \times 256$&$256 \times 256$ \\
    \bottomrule
    \end{tabular}
    }
    \label{tab:suppl_data}
\end{table*}


\textbf{Datasets.} We conduct experiments on three common-sense CD datasets, i.e., VL-CMU-CD~\cite{alcantarilla2018street}, PCD~\cite{sakurada2015change}, and CDNet~\cite{goyette2012changedetection}, and three remote sensing CD datasets, \ie, LEVIR-CD~\cite{chen2020spatial}, CDD~\cite{lebedev2018change}, and AICD~\cite{bourdis2011constrained}. Details of each dataset are shown in Table~\ref{tab:suppl_data}.
For VL-CMU-CD~\cite{alcantarilla2018street}, we randomly sample 80\% image pairs for training and 20\% for testing. The training images are augmented by randomly clipping and flipping.
For PCD~\cite{sakurada2015change}, we crop the image with a size of $224\times 224$ and partition the training and testing set with a ratio of $8:2$. The training images are upsampled to $320\times320$ with flipping during the training phase.
For CDNet~\cite{goyette2012changedetection}, we use the training and testing dataset as proposed in~\cite{huang2020change}. 
For LEVIR-CD~\cite{chen2020spatial}, we crop images into small patches of size $256 \times 256$ without overlapping and augment training images with rotation and color transformation. 
%
For CDD~\cite{lebedev2018change}, we adopt the default dataset partition to train and evaluate various change detectors. 
For AICD~\cite{bourdis2011constrained}, we crop image pairs randomly and augment the training images by rotation and color vibration. 

\textbf{Evaluation metrics.} Following \cite{chen2021a}, we use precision (P), recall (R), F1-measure (F1), overall accuracy (OA) and intersection over union (IoU) to evaluate different CD methods, which can be calculated as:
\begin{equation}
    \begin{aligned}
    P &= \frac{TP}{TP + FP}, \\
    R &= \frac{TP}{TP + FN}, \\
    F1 &= \frac{2\times P \times R}{P + R}, \nonumber
    \end{aligned}
\end{equation}
\begin{equation}
    \begin{aligned}
        OA &= \frac{TP + TN}{TP+TN+FP+FN}, \\
        IoU &= \frac{TP}{TP+FP+FN}.
    \end{aligned}
\end{equation}
where $TP$, $TN$, $FP$ and $FN$, represent true positive, true negative, false positive, false negative, respectively.

\begin{figure*}[t]  
\centering  
\includegraphics[width=\linewidth]{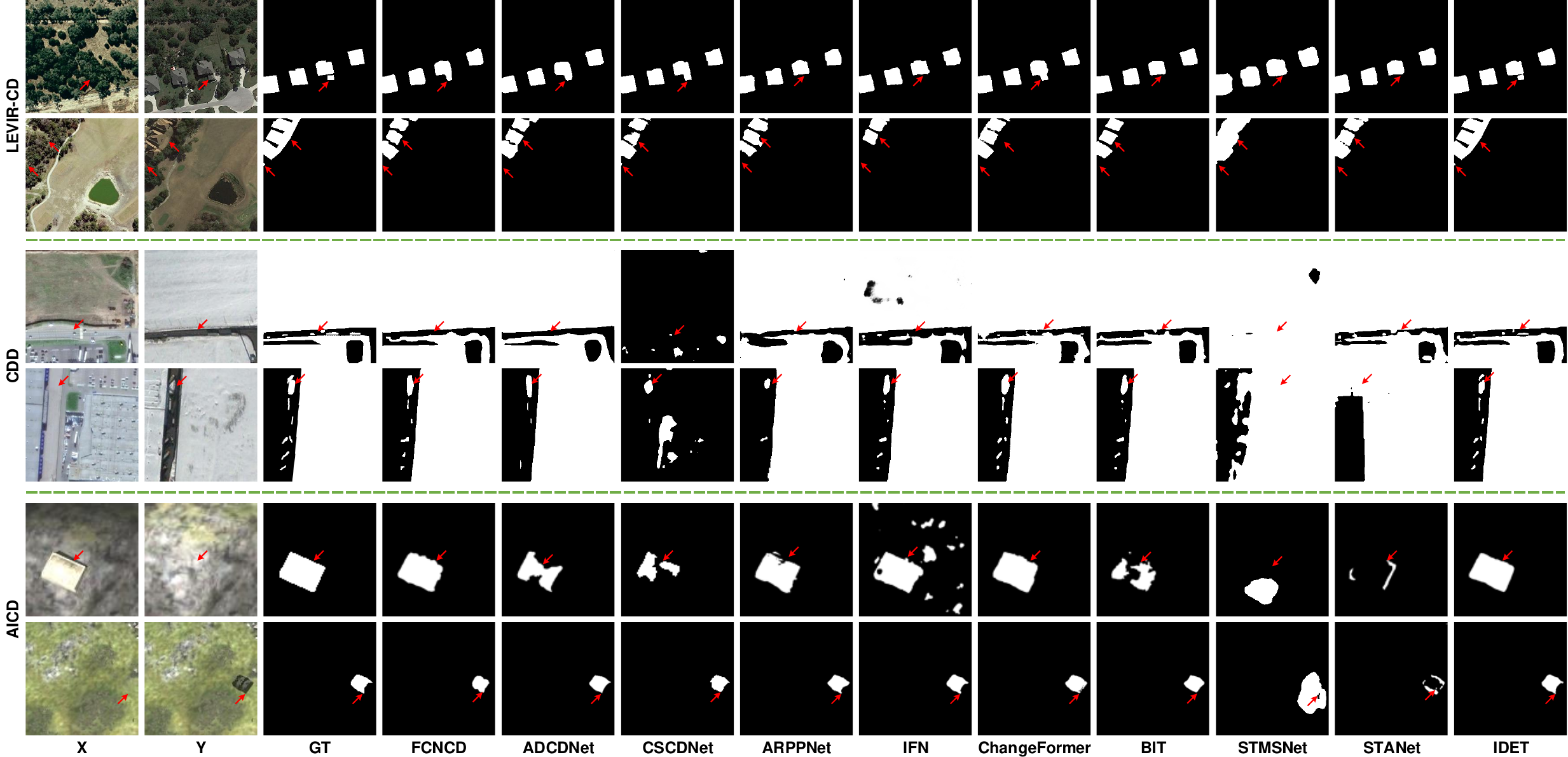}  
\caption{Detection results of different CD methods on LEVIR-CD~\cite{chen2020spatial}, CDD~\cite{Shunping2019Fully} and AICD~\cite{bourdis2011constrained} datasets.}  
\label{fig:remotes}  
\end{figure*}

\begin{table*}[h]
    \centering
    \caption{ Quantitative comparison of CD methods on VL-CMU-CD, PCD and CDNet. The best results are marked with \first{orangered} and the second-best are marked with \second{purple}.}
    \resizebox{0.95\linewidth}{!}{
    \begin{tabular}{l|ccccc|ccccc|ccccc}
    \toprule
    \multirow{2}{*}{Method}&\multicolumn{5}{c|}{VL-CMU-CD} & \multicolumn{5}{c|}{PCD} & \multicolumn{5}{c}{CDNet}\\ 
    
    &P & R & F1  & OA & IoU &P& R & F1 &  OA & IoU &P& R & F1 & OA & IoU \\
    \midrule

    FCNCD & 84.2& 84.3& 84.2  & 98.4 & 72.9 &68.7&61.7& 65.0 & \first{94.1} & 48.0 &82.9&88.3& 85.5  & 99.2 & 76.0 \\
    
    ADCDNet & 92.8& 94.3& \second{93.5}  & \second{99.3} & \second{87.9} &\first{78.8}&73.4& \first{76.0}  & \second{90.0} & \first{62.3} &89.0&85.5& 87.2 & \second{99.3} & 77.9 \\
    
    CSCDNet &89.2&91.1& 90.1&98.7&82.1 &66.0&72.6& 69.1 & 83.7 & 51.9 &\second{93.9}&82.3& 87.7 & 99.0 & 78.4 \\

    ARPPNet&88.2&85.5&86.8&98.4&76.8&70.8&61.5&65.8&83.7&49.5&83.7&73.6&78.3&93.3&63.8 \\

    IFN &\second{93.1}& 75.9& 83.6 & 98.1 & 71.4 &69.5&75.7& \second{72.5} & 87.2 & 57.7 &93.1&80.0& 86.1 & 98.9 & 75.3  \\
    
    BIT & 90.5& 87.6& 89.0  & 98.8 & 80.6 &\second{77.7}&57.3& 66.0  & 85.4 & 50.1 &\first{95.4} &81.9&88.1 &99.1 &78.9  \\

    STANet &73.1&\first{95.8} & 82.9  & 98.0 & 70.7 &69.6&52.7& 60.0 & 79.9 & 41.2 &72.2& \first{94.3}& 81.8  &98.3  &69.2  \\ 
    
    ChangeFormer &88.0 &88.6 &88.3 &98.7 &79.2 &74.0 & 69.2 & 71.5 & 86.6 & 56.4 &93.4 &85.9 &\second{89.5} &99.2 &\second{81.3} \\
    
    STMSNet&38.2&69.4&49.3&87.3&27.4&60.0&\first{79.5}&68.4&76.1&50.2&68.4&80.9&74.1&92.6&57.3 \\

    SFBNet&43.3&46.6&44.9&64.9&37.9 &66.3&72.9&69.4&83.4&53.5 & 91.6&85.9&88.7&99.2&80.4\\

    RCTNet&79.2&88.7&83.7&98.2&72.2 &75.8& 67.1&71.2&87.9&56.9 &92.1&83.6&87.6&99.2&79.6\\
    IDET~(Ours) &\first{93.5}&\second{94.5}& \first{94.0}  &\first{99.4} & \first{88.7} &74.2&\second{77.9}& \first{76.0} & 88.8 & \second{61.9} &90.8&\second{94.1} & \first{92.4} & \first{99.7} & \first{86.2} \\
    \bottomrule
    \end{tabular}}
    
    \label{tab:my_commons}
\end{table*}

\begin{table*}[h]
    \centering
    \small
    \caption{Quantitative comparison of CD methods on LEVIR-CD, CDD, and AICD. The best results are marked with \first{orangered} and the second-best are marked with \second{purple}.}
    %
     \resizebox{0.95\linewidth}{!}{
    \begin{tabular}{l|ccccc|ccccc|ccccc}
    \toprule
    \multirow{2}{*}{Method}&\multicolumn{5}{c|}{LEVIR-CD} & \multicolumn{5}{c|}{CDD} & \multicolumn{5}{c}{AICD}\\ 
    &P& R & F1  & OA & IoU &P& R & F1 &  OA & IoU &P& R & F1 & OA & IoU \\
    \midrule

    FCNCD &89.7&84.0& 86.8 & 97.7 & 77.1 &85.1&69.8& 76.7 & 97.5 & 63.7 &81.4&87.7& 84.4 & 99.9 & 72.7\\ 

    ADCDNet &90.4&81.8& 85.9 & 97.4 & 75.4 &92.5&78.8& 85.1 & 98.5 & 75.1 &85.0&\second{99.0}& \second{91.5}  & 99.9 & 84.2 \\
    CSCDNet &84.6&81.6& 83.1&96.8 & 71.5 &79.1&48.9& 60.4 & 90.9 & 42.7 &92.5&88.9& 90.7 & 99.9 & 82.6 \\

    ARPPNet&77.9&72.7&75.2&97.0&64.1&72.8&51.5&60.3&96.9&47.1&78.9&97.6&87.3&99.9&78.0 \\

    IFN &89.7&79.8& 84.5 & 96.9 & 72.7 &\first{94.2}&69.2& 79.8 & 97.7 & 66.5 &84.6&97.7& 90.7 & 99.9 & 82.9 \\
    
    BIT &\first{92.1}&81.6& 86.5  & 97.7 & 76.3 &93.5&80.2& 86.3 & 98.7 & 76.6 &\second{94.2}&92.4& \first{93.2} & 99.9 & \first{87.4}  \\ 
    
    STANet &84.8&69.9& 76.6 & 95.3 & 61.8 &71.6&69.6& 70.6 & 95.0 & 54.6 &46.0&35.6& 40.1 & 99.4 & 22.4\\ 
    ChangeFormer&90.7&85.5&\second{88.0}&\second{98.0}&\second{79.0}&89.2&73.0&80.3&98.0&68.0&82.7&98.3&89.8&99.9&81.5 \\
   
    STMSNet&54.0&\first{88.2}&67.2&92.7&51.7&40.4&70.3&51.3&88.9&35.1&28.7&\first{99.3}&44.5&99.2&28.7 \\
    
    SFBNet &83.2 &69.4 &75.7 &97.3 &65.0 & \second{93.6}&\second{80.4} &\second{86.5} &\first{99.0} &\first{78.5} &76.1&93.3&83.8&99.9&72.6 \\
    
    RCTNet&81.4&74.5&77.8&97.7&68.8&87.6&73.8&80.1&98.4&70.1   &79.5&96.0&87.0&99.9&77.3\\
    
    IDET~(Ours) & \second{91.3}& \second{86.6} & \first{88.9} &  \first{98.1} & \first{80.2} & 91.1& \first{83.1}& \first{86.9} & \second{98.7} & \second{77.6} & \first{96.2}& 90.4& \first{93.2} & \first{99.9} & \second{87.2} \\

    \bottomrule
    \end{tabular}}
    \label{tab:my_remotes}
\end{table*}

\subsection{Comparison Result}
\label{subsec:compar}
\figref{fig:commons} shows some representative CD examples of the state-of-the-art methods on VL-CMU-CD, PCD, and CDNet datasets. 
FCNCD~\cite{long2015fully} only captures the main parts of changes, which loses the details of the changes as shown in the 1st to 4th rows of Fig.~\ref{fig:commons}. As illustrated in the 2nd and 3rd rows of \figref{fig:commons}, CSCDNet~\cite{sakurada2020weakly} is fail to detect changes in lighting difference scenes. IFN~\cite{zhang2020deeply} also loses the detailed information of changes, even losing the change object in the 2nd row. Transformer-based methods ChangeFormer~\cite{bandara2022transformer} and BIT~\cite{chen2021a} are all easily affected by shadows and background clusters, which results in false detection in the 2nd and 4th rows. STANet~\cite{chen2020spatial} generates incomplete CD results as shown in the 4th and 5th rows of Fig.~\ref{fig:commons}. Among the compared methods, ADCDNet~\cite{huang2020change} gains good CD results on all three CD datasets. However, we can find that the results of IDET are more complete and have more details as shown in the last column of Fig.~\ref{fig:commons}.

\begin{table*}[ht]
    \centering
    \caption{ Quantitative results of SOTA CD methods with and without IDET on the VL-CMU-CD, PCD, and CDNet datasets.}
    \resizebox{0.8\linewidth}{!}{
    \begin{tabular}{lllllll}
    \toprule
        
    Dataset&Method&P  & R  & F1  & OA  & IoU  \\
    \midrule

    \multirow{8}{*}{VL-CMU-CD}
    &\cellcolor{tabgray}ADCDNet & \cellcolor{tabgray}92.8&\cellcolor{tabgray}94.3&\cellcolor{tabgray}93.5&\cellcolor{tabgray}99.3&\cellcolor{tabgray}87.9 \\
    &\cellcolor{tabgray}\textbf{ADCDNet with IDET}&\cellcolor{tabgray}\textbf{\textbf{92.5 (-0.3)}} &\cellcolor{tabgray}\textbf{95.4 (+1.1)} &\cellcolor{tabgray}\textbf{93.9 (+0.4)} &\cellcolor{tabgray}\textbf{99.3 (+0.0)} &\cellcolor{tabgray}\textbf{88.6 (+0.7)} \\

    &BIT&90.5&87.6&89.0&98.8&80.6\\
    &\textbf{BIT with IDET}& \textbf{92.4 (+1.9)}&\textbf{88.3 (+0.7)} &\textbf{90.3 (+1.3)} &\textbf{99.0 (+0.2)} &\textbf{82.7 (+2.1)} \\

    &\cellcolor{tabgray}STANet&\cellcolor{tabgray}73.1&\cellcolor{tabgray}95.8&\cellcolor{tabgray}82.9&\cellcolor{tabgray}98.0&\cellcolor{tabgray}70.7 \\
    &\cellcolor{tabgray}\textbf{STANet with IDET} &\cellcolor{tabgray}\textbf{82.2 (+9.1)} &\cellcolor{tabgray}\textbf{95.8 (+0.0)} &\cellcolor{tabgray}\textbf{88.5 (+5.6)} &\cellcolor{tabgray}\textbf{98.6 (+0.6)}&\cellcolor{tabgray}\textbf{79.3 (+8.6)}\\

    &ChangeFormer & 88.0&88.6&88.3&98.7&79.2 \\
    &\textbf{ChangeFormer with IDET}&\textbf{89.7 (+1.7)}&\textbf{89.6 (+1.0)}&\textbf{89.6 (+1.3)}&\textbf{98.8 (+0.1)}&\textbf{81.8 (+2.6)}\\

    \midrule
    \multirow{8}{*}{PCD}
    &\cellcolor{tabgray}ADCDNet&\cellcolor{tabgray}78.8&\cellcolor{tabgray}73.4&\cellcolor{tabgray}76.0&\cellcolor{tabgray}90.0&\cellcolor{tabgray}62.3\\
    &\cellcolor{tabgray}\textbf{ADCDNet with IDET}&\cellcolor{tabgray}\textbf{79.8 (+1.0)} &\cellcolor{tabgray}\textbf{73.0 (-0.4)} &\cellcolor{tabgray}\textbf{76.3 (+0.3)} &\cellcolor{tabgray}\textbf{90.2 (+0.2)} &\cellcolor{tabgray}\textbf{62.5 (+0.2)}\\
        
    &BIT&77.7&57.3&66.0&85.4&50.1 \\
    &\textbf{BIT with IDET}&\textbf{77.5 (-0.2)} &\textbf{58.8 (+1.5)} &\textbf{66.5 (+0.5)} &\textbf{85.7 (+0.3)} &\textbf{51.1 (+1.0)} \\

    &\cellcolor{tabgray}STANet&\cellcolor{tabgray}69.6&\cellcolor{tabgray}52.7&\cellcolor{tabgray}60.0&\cellcolor{tabgray}79.9&\cellcolor{tabgray}41.2 \\
    &\cellcolor{tabgray}\textbf{STANet with IDET}&\cellcolor{tabgray}\textbf{58.2 (-11.4)} &\cellcolor{tabgray}\textbf{82.5 (+29.8)} &\cellcolor{tabgray}\textbf{68.2 (+8.2)}&\cellcolor{tabgray}78.5 (-1.4) &\cellcolor{tabgray}\textbf{51.3 (+10.1)}\\

    &ChangeFormer&74.0&69.2&71.5&86.6&56.4 \\
    &\textbf{ChangeFormer with IDET} &\textbf{73.4 (-0.6)} &\textbf{70.1(+0.9)} &\textbf{71.7 (+0.2)} &\textbf{86.7 (+0.1)}&\textbf{56.8 (+0.4)}\\

    \midrule
    \multirow{8}{*}{CDNet}
     &\cellcolor{tabgray}ADCDNet&\cellcolor{tabgray}89.0&\cellcolor{tabgray}85.5&\cellcolor{tabgray}87.2&\cellcolor{tabgray}99.3&\cellcolor{tabgray}77.9 \\
    &\cellcolor{tabgray}\textbf{ADCDNet with IDET}&\cellcolor{tabgray}\textbf{91.6 (+2.6)} &\cellcolor{tabgray}\textbf{83.5 (-2.0)} &\cellcolor{tabgray}\textbf{87.4 (+0.2)} &\cellcolor{tabgray}\textbf{99.4 (+0.1) }&\cellcolor{tabgray}\textbf{80.0 (+0.1)}  \\
    
    &BIT&95.4&81.9&88.1&99.1&78.9 \\
    &\textbf{BIT with IDET}&\textbf{95.6 (+0.2)} &\textbf{81.9 (+0.0) }&\textbf{88.3 (+0.2)} &\textbf{99.2 (+0.1)} &\textbf{79.3 (+0.4)} \\

    &\cellcolor{tabgray}STANet&\cellcolor{tabgray}72.2&\cellcolor{tabgray}94.3&\cellcolor{tabgray}81.8&\cellcolor{tabgray}98.3&\cellcolor{tabgray}69.2 \\
    &\cellcolor{tabgray}\textbf{STANet with IDET}&\cellcolor{tabgray}\textbf{84.7 (+12.5)} &\cellcolor{tabgray}\textbf{89.9 (-4.4)} &\cellcolor{tabgray}\textbf{87.2 (+5.4)} &\cellcolor{tabgray}\textbf{98.8 (+0.5)} &\cellcolor{tabgray}\textbf{77.9 (+8.7)} \\

    &ChangeFormer&93.4&85.9&89.5&99.2&81.3 \\
    &\textbf{ChangeFormer with IDET}&\textbf{93.7 (+0.3)}&\textbf{85.8 (-0.1)}&\textbf{89.6 (+0.1)}&\textbf{99.3 (+0.1)}&\textbf{81.5 (+0.2)}\\

    \bottomrule
    \end{tabular}}
    \label{tab:with_idet}

\end{table*}

Fig.~\ref{fig:remotes} shows some typical detection results of different CD methods on LEVIR-CD, CDD and AICD datasets. IFN \cite{zhang2020deeply}, BIT \cite{chen2021a} and STANet \cite{chen2020spatial} fail to detect the small house in the 1st row of Fig.~\ref{fig:remotes}. In the 2nd row, only IDET can detect the connected changes.
CSCDNet \cite{sakurada2020weakly} achieves the worst result on the CDD dataset, in the 3rd and 4th rows of \figref{fig:remotes}, which demonstrates that CSCDNet cannot deal with the large changes.
ADCDNet \cite{huang2020change} is good at detecting the changes of the image pairs of common sense datasets, but failed to capture the integral changes of remote sensing datasets in the 4th and 5th rows of \figref{fig:remotes}.
IFN \cite{zhang2020deeply} is easily affected by season variation and lighting difference, resulting bad detection in the 3rd and 5th rows of \figref{fig:remotes}.
ChangeFormer \cite{bandara2022transformer} fails to detect the changes in fine-grained level in the 1st and 2nd rows of \figref{fig:remotes}. 
BIT \cite{chen2021a} is easily affected by shadows and lighting differences, leading to false detection in the 1st and 5th rows of \figref{fig:remotes}.
STANet \cite{chen2020spatial} has bad detection results (see the 2nd, 5th and 6th rows of \figref{fig:remotes}) when there exist large lighting differences. 
In the last two image pairs of Fig.~\ref{fig:remotes}, we find that most of the compared methods are easily affected by overexposure. On the contrary, only IDET can generate promising CD results.

\tableref{tab:my_commons} shows the quantitative comparison on common sense datasets. 
On VL-CMU-CD, almost all the criteria of IDET are better than other compared methods.
Take the ADCDNet~\cite{huang2020change} for example, whose F1, OA, and IoU are very close to IDET. Only 0.5\% and 0.8\% improvements are achieved by IDET on F1 and IoU, respectively. Compared with the best remote sensing CD method BIT~\cite{chen2021a}, IDET achieves $5.6\%$, $0.6\%$, and $10.0\%$ relative F1, OA, and IoU improvements, respectively. The F1, OA, and IoU of ChangeFormer are 88.3\%, 98.7\%, and 79.2\% respectively, which are lower than IDET by 5.7\%, 0.7\%, and 9.5\%, respectively.
The F1 and IoU of STMSNet are 49.3\% and 27.4\%, fewer than IDET by 44.7\% and 61.3\%. STMSNet performs poorly because the employed naive transformer can't overcome the large camera variations and lighting differences. And the fixed binary threshold doesn't balance the precision (38.1\%) and recall (69.4\%) well. SFBNet~\cite{sun2024spatial} achieves the F1 by 44.9\%, which is greatly lower than IDET due to lacking optimization about the feature difference.
On the PCD dataset, there are only four CD methods, whose F1 values are higher than 70\%. ADCDNet~\cite{huang2020change} shows the best performance than other CD methods. Compared with ADCDNet~\cite{huang2020change}, the IoU of IDET are a bit lower than ADCDNet~\cite{huang2020change}. Taking the IFN for example, IDET improves the F1, OA, and Iou from 72.5\% to 76.0\%, from 87.2\% to 88.8, and from 57.7\% to 61.9, respectively. RCTNet~\cite{gao2024relating} achieves a F1 and IoU by 71.2\%and 56.9\%, respectively. It adopts a multi-scale fusion mechanism to refine the predicted maps that show superiority than the SFBNet. 
On the CDNet dataset, only the F1 value and IoU of IDET are higher than 90\% and 85\%, respectively. Compared with ARPPNet, IDET achieves 14.1\%, 6.4\%, and 22.4\% improvements in terms of F1, OA, and IoU, respectively. ChangeFormer~\cite{bandara2022transformer} achieves the best CD performance than other comparison methods. However, compared with ChangeFormer, IDET achieves $3.2\%$, $0.5\%$, and $6.0\%$ relative F1, OA, and IoU improvements, respectively. SFBNet achieves the F1 by 88.7, which is lower than our IDET by 3.7\%. Compared with RCTNet, SFBNet predicts change maps only from high-level semantic features that show higher recall~(R). While RCTNet fuses multi-scale differences for final CD maps, achieving the better precision~(P) of 92.1\%.

Table~\ref{tab:my_remotes} shows the quantitative comparison of different CD methods on remote sensing datasets. We can find that IDET has excellent performances on three remote sensing datasets in most criteria. Concretely, on LEVIR-CD, IDET achieves $1.0\%$, $0.1\%$ and $1.5\%$ relative F1, OA and IoU improvements than ChangeFormer~\cite{bandara2022transformer}, respectively.
The F1, OA, and IoU of ADCDNet are 85.9\%, 97.4\%, and 75.4\%, which are lower than IDET of 3\%, 0.7\%, and 4.8\%, respectively. ARPPNet obtains the lowest F1~(75.2) on the LEVIR-CD dataset. Although adopting the larger convolution fields to capture the difference, the early feature concatenation leads to the poor quality of change-related features.
On CDD dataset, SFBNet~\cite{sun2024spatial} is the best CD method among the comparison CD methods. The R and F1 of IDET are higher than SFBNet by $2.7\%$ and $0.4\%$, respectively. Compared with another Transformer-based method, BIT~\cite{chen2021a}, IDET achieves 1.1\% and 1.0\% improvements on F1 and IoU, respectively. 
Compared with another CNN-Transformer-based method, RCTNet~\cite{gao2024relating}, IDET achieves 6.9\%, 0.3\%, and 7.5\% improvements in terms of F1, OA, and IoU, respectively. 
On AICD, IDET has a similar performance to BIT, which has the same F1 values (93.2\%) and OA (99.9\%). Compared with STMSNet, IDET improves the F1 value from 44.5\% to 93.2\%.
Except for STANet and STMSNet, the F1 values and IoU of all other CD methods are higher than 80\% and 70\%, respectively.
STANet employs the fixed threshold ($\theta$) to binarize the absolute feature differences to produce the change mask. The higher $\theta$ leads to loss of object changes easily, which makes STANet has the lowest recall~(R) among the nine mentioned methods on AICD.

\begin{figure*}[t]
\centering 
\includegraphics[width=0.9\linewidth]{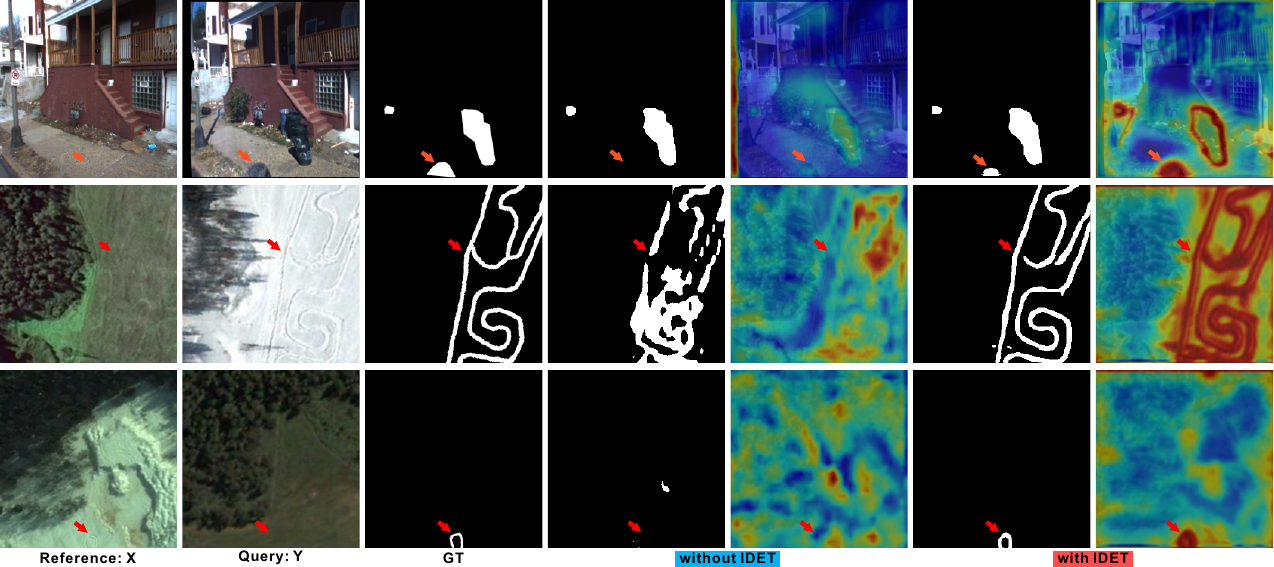}
\caption{Three cases with/without IDET. The main advantages are highlighted via red arrows.}\label{fig:rebuttal1}
\label{rebuttalfig:vis}
\end{figure*}

\begin{table}[t]
    \centering
      \caption{Ablation studies on multi scales and three transformers $\tau_{\text{ref}}(\cdot)$\&$\tau_{\text{que}}(\cdot)$, $\tau_{\text{diff}}(\cdot)$~(3rd transformer). We use the convolution block (consists of four convolution layers with batch normalization and ReLU) to replace the IDET and get a variant (\ie, Multi-Scale-CNN-feat.~diff.~enhance.).
    We make this experiment on VL-CMU-CD. The \textbf{bold} is the best.
    } 
    \begin{tabular}{lccc}
    \toprule
    Methods & F1 & OA & Iou \\
    \midrule
    w/o feat.~diff.~enhance.  
                    & 93.2 & 99.3 & 87.7 \\

    Multi-Scale-CNN-feat.~diff.~enhance. & 93.6 & 99.3 & 88.0\\

    {\bf Multi-Scale.IDET-feat.~diff.~enhance. (Ours)} & {\bf94.0} & {\bf99.4} & {\bf88.7} \\ 
    Naive-Multi-Scale.IDET-feat.~diff.~enhance.
                    & 93.8 & 99.3 & 88.4 \\ 
    Single-Scale.IDET-feat.~diff.~enhance.
                    & 93.7 & 99.3 & 88.2 \\ 
    \midrule
    Multi-Scale.IDET~w/o~$\tau_{\text{diff}}(\cdot)$   
                    & 93.5 & 99.3 & 87.9 \\  
    Single-Scale.IDET~w/o~$\tau_{\text{diff}}(\cdot)$
                    & 93.3 & 99.3 & 87.5 \\ 
    \midrule
    Multi-Scale.IDET~w/o~$\tau_{\text{ref}}(\cdot)$\&$\tau_{\text{que}}(\cdot)$
                    & 92.9 & 99.3 & 86.7 \\ 
    Single-Scale.IDET~w/o~$\tau_{\text{ref}}(\cdot)$\&$\tau_{\text{que}}(\cdot)$
                    & 92.4 & 99.2 & 86.0 \\ 
    \bottomrule
    \end{tabular}
    \label{rebuttaltab:ablation}
\end{table}

\subsection{Extension to Existing CD Methods}
\label{subsec:extension}

IDET aims to extract long-range information from bi-temporal images to improve the quality of feature differences, which can be used in other existing CD methods to enhance their detection accuracy. 
To demonstrate this, we equip two CNN-based methods (\ie, ADCDNet \cite{huang2020change} and STANet \cite{chen2020spatial} and two transformer-based methods (\ie, BIT \cite{chen2021a} ChangeFormer \cite{bandara2022transformer}) with our IDET. Specifically, we use the backbones of the four methods to extract the features of two input images (\ie, $\mathbf{R}_\text{x}$ and $\mathbf{R}_\text{y}$). Then, we feed the extracted features to our IDET introduced in \secref{subsec:idet} and map the outputs to the changing mask.
Note that, ChangeFormer concatenates the bi-temporal features to predict the change regions instead of calculating the feature difference, we replace its feature concatenation MLP with our IDET to calculate the change-related features. 
\tableref{tab:with_idet} shows the results of each baseline with and without our IDET on three datasets (\ie, VL-CMU-CD, PCD, and CDNet ), where the improvements are shown in the brackets. We find that baselines with IDET can achieve better F1  scores on all datasets than the initial counterparts. In particular, the F1 score of STANet has significantly improved, showing an absolute increase of 5.6, 8.2, and 5.4 on the three datasets.
It demonstrates that the proposed IDET can be embedded into other CD methods to improve detection performance.
This performance may be improved when we enhance robustness the IDET.

\subsection{Discussion}
\label{subsec:astudy}

\subsubsection{\textbf{Differences to existing transformers}} %
In this work, for the first attempt, explore the critical role of the feature difference quality to the CD (See Fig.~\ref{fig:analysis}).
The intuitive idea of IDET is to extract difference information for the representations of input images (\ie, $\mathbf{R}_\text{x}$ and $\mathbf{R}_\text{y}$) and use them to guide the enhancement of the feature difference $\mathbf{D}$.
To this end, we propose a novel transformer-based module that is different from previous ones (See Fig.~\ref{fig:msidet}). 
\textit{First}, the whole IDET is novel but not a naive combination of three transformers. The transformers $\tau_\text{ref}$ and $\tau_\text{que}$ are to extract difference-related information in corresponding features (\ie, $\mathbf{R}_\text{x}$ and $\mathbf{R}_\text{y}$), and we design an extra function $\text{diff}(\cdot)$ to further refine the semantic differences.
\textit{Second}, the transformer $\tau_\text{diff}$ is a dynamic transformer and basically different from existing ones while it allows borrowing the extra information (\ie, $\Tilde{\mathbf{D}}$) to guide the updating of the input $\mathbf{D}$ dynamically (See Eq.~(\ref{eq:idet_former3})). 
\textit{Third}, IDET allows the iterative optimization that benefits the detection accuracy significantly as demonstrated in \tableref{rebuttaltab:ablation}. By replacing our IDET by a CNN model, we can see that F1 drops from 94.0 to 93.6. Also, $\tau_{\text{ref}}(\cdot)$, $\tau_{\text{que}}(\cdot)$ and $\tau_{\text{diff}}(\cdot)$ are replaced by a convolution layer to study each function in our IDET, which shows that $\tau_{\text{ref}}(\cdot)$ and $\tau_{\text{que}}(\cdot)$ benefit the CD performance more than $\tau_{\text{diff}}(\cdot)$.

\subsubsection{\textbf{Visualization of the feature difference enhancement}}
\begin{figure}[htb]  
\centering  
\includegraphics[width=\linewidth]{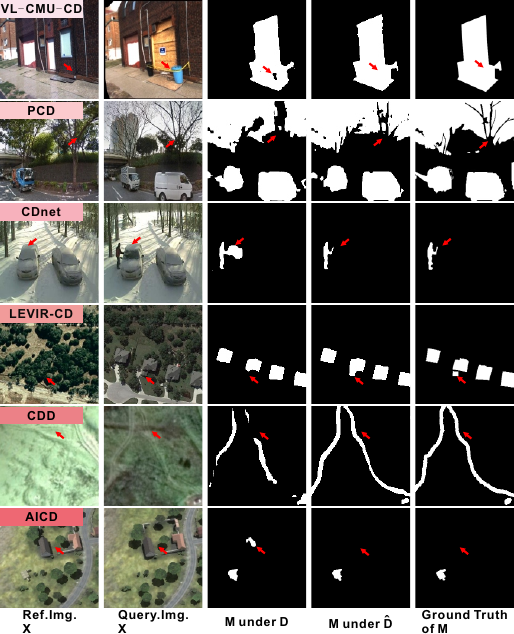} 
\caption{Comparison of $\textbf{M}$ under $\textbf{D}$ with $\textbf{M}$ under $\hat{\textbf{D}}$ on VL-CMU-CD, PCD, CDNet, LEVIR-CD, CDD, and AICD.
}  
\label{fig:suppl_diff}  
\end{figure}

We have visualized the importance of the feature difference enhancement in Fig.~\ref{fig:analysis}. Here, in Fig.~\ref{rebuttalfig:vis}, we further present three cases to compare the feature difference before and after enhancement ($5$th column~vs.~$7$th column) and detection results with/without IDET. Clearly, IDET enhances the desired changes in the feature difference map, leading to obvious advantages in final evaluations. \figref{fig:suppl_diff} shows the detailed comparison of $\mathbf{M}$ under $\mathbf{D}$ and $\hat{\mathbf{D}}$ on six datasets, which demonstrates that IDET improves the final results effectively.

\begin{table*}[]
    \centering
    \caption{Results of different $T$ in IDET.}
    \begin{tabular}{ccccccccccccccccccc}
    \toprule
        Dataset &  \multicolumn{3}{c}{CMU} & \multicolumn{3}{c}{PCD} & \multicolumn{3}{c}{CDNet} & \multicolumn{3}{c}{LEVIR} & \multicolumn{3}{c}{CDD} & \multicolumn{3}{c}{AICD} \\
    \cmidrule(r){1-1} \cmidrule(r){2-4} \cmidrule(r){5-7} \cmidrule(r){8-10} \cmidrule(r){11-13}
    \cmidrule(r){14-16} \cmidrule(r){17-19} 
    Iter & F1&OA&IoU& F1&OA&IoU& F1&OA&IoU& F1&OA&IoU& F1&OA&IoU& F1&OA&IoU \\
    \midrule
        0 &93.2&99.3&87.8&71.2&88.3&58.5&91.7&99.6&85.7&87.0&98.0&79.3&80.0&98.1&69.4&89.5&99.9&81.4 \\
        1 &93.6&93.8&88.4&73.8&89.5&61.7&92.6&99.7&87.0&82.9&97.3&74.0&82.9&98.5&73.4&90.5&99.9&82.8 \\
        2 &94.0&99.4&88.7&76.0&88.8&61.9&92.4&99.7&86.2&88.9&98.1&80.2&85.4&98.5&78.2&91.4&99.9&84.4 \\
        3 &94.0&99.4&88.6&73.9&88.4&59.3&93.8&99.6&88.8&61.7&96.1&50.4&83.0&98.7&73.6&86.9&99.9&76.8 \\
        4 &93.9&99.4&88.6&74.1&88.5&59.4&93.7&99.6&88.7&83.2&98.2&75.1&83.5&98.7&74.1&89.5&99.9&81.4 \\
        5 &93.9&99.3&88.6&74.6&88.2&59.4&93.8&99.7&88.9&83.0&98.2&74.8&83.0&98.7&73.3&90.1&99.9&82.0 \\
    \bottomrule
    \end{tabular}

     \label{tab:tab_IDET}
\end{table*}

\subsubsection{\textbf{Effectiveness of multiple losses with $\mathcal{F}$ and $\mathcal{CE}$}}
We add supervision on all predicted change maps of $\mathbf{M}^0,\ldots,\mathbf{M}^L$ and $\mathbf{M}$ to train our model. To validate the effectiveness of using multiple supervision, we retrain our network only using single supervision on the final change map $\mathbf{M}$ with Cross entropy loss~($\mathcal{SS} + \mathcal{CE}$), multiple supervision with Focal loss~($\mathcal{MS} + \mathcal{F}$), and multiple supervision with Cross entropy loss~($\mathcal{MS} + \mathcal{CE}$). As shown in Table~\ref{tab:my_abla_sce}, we find that \ding{182} using multiple supervisions~($\mathcal{MS}+ \mathcal{CE}$) achieves consistently F1, OA and IoU improvements over using single supervision~($\mathcal{SS} + \mathcal{CE}$); \ding{183} using Cross entropy loss~($\mathcal{MS} + \mathcal{CE}$) is more suitable for CD than using Focal loss~($\mathcal{MS} + \mathcal{F}$).

\begin{table}[t]
    \centering
    \caption{Results of IDET with single supervision~($\mathcal{SS}$) on the final layer and multiple supervision~($\mathcal{MS}$), Focal loss~($\mathcal{F}$) and Cross entropy loss~($\mathcal{CE}$). The \textbf{bold} is the best.}
     \resizebox{\linewidth}{!}{
    \begin{tabular}{l|ccc|ccc|ccc}
    \toprule
    \multirow{2}{*}{Dataset}&\multicolumn{3}{c|}{$\mathcal{SS}$+$\mathcal{C}$E} & \multicolumn{3}{c|}{$\mathcal{MS}$+$\mathcal{F}$} & \multicolumn{3}{c}{$\mathcal{MS}$+$\mathcal{CE}$}\\ 

    & F1  & OA & IoU  & F1 &  OA & IoU & F1 & OA & IoU \\
    \midrule
    VL-CMU-CD &91.6 &99.3 &87.3  &91.6 &99.2 &84.6  &\textbf{94.0} &\textbf{99.4} &\textbf{88.7}\\
    
    PCD & 75.0 &\textbf{88.9} &60.3 & 70.1 &88.1 &54.4 &\textbf{76.0} &88.8 &\textbf{61.9}\\

    CDNet &\textbf{92.4} &99.6 &86.1 &89.9 &99.5 &81.6 &\textbf{92.4} &\textbf{99.7} &\textbf{86.2}\\

    LEVIR-CD &87.8 &97.8 &78.3  &82.5 &97.0 &70.3 &\textbf{88.9} &\textbf{98.1} &\textbf{80.2}\\

    CDD & 78.0 &97.6 &64.3 &70.8 &97.1 &55.1 &\textbf{85.4} &\textbf{98.5} &\textbf{75.2}\\

    AICD &88.5 &\textbf{99.9} &79.4  &88.8 &\textbf{99.9} &80.4 &\textbf{91.4}  &99.3 &\textbf{82.5}\\
    \bottomrule
    \end{tabular}}
    \label{tab:my_abla_sce}
\end{table}

\subsubsection{\textbf{Iteration of IDET}}
IDET iteratively enhances the change-related features to highlight the change regions and suppress background areas.
We set different iteration numbers~($T$) for IDET to study the influence of $T$. Table~\ref{tab:tab_IDET} shows the results of IDET with different iterations on the adopted six datasets. Deviating from an iteration number of $T=2$, whether increasing or decreasing the value, does not enable IDET to reach its peak performance. This is because a higher iteration number tends to over-suppress the changed regions, whereas a lower number fails to adequately filter out background noises.
Specifically, our IDET achieves the best performances on each dataset except CDNet when $T=2$.
Because the CDNet is a \textit{video frame-based} change detection dataset with a greater number of images, where bi-temporal images contain more similar backgrounds that can be modeled by a deeper IDET. However, the other five datasets have fewer images, and each image pair shows distinctive background details. Therefore, an iteration number of $T=2$ makes IDET achieve an optimal balance between enhancing the changed regions and dampening background noises, thus yielding the best overall performance on most datasets.

\begin{figure*}[htb]
\centering
\includegraphics[width=0.9\linewidth]{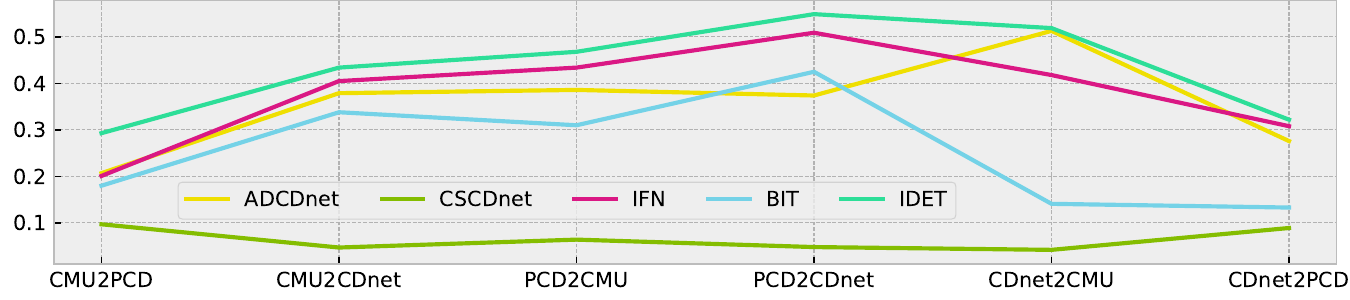}
\caption{Cross validation of comparison methods (\ie ADCDNet, CSCDNet, IFN, BIT) on VL-CMU-CD, PCD, and CDNet.}\label{fig:rebuttal2}
\label{rebuttalfig:cross}
\end{figure*}

\begin{table}[t]
    \centering
      \caption{Comparison on model efficiency. Running time (RT) and GFLOPs of IDET with different $T$ and comparison methods. We set a CNN model to replace the IDET and let it has even larger GFLOPS than the IDET~(T=3). The input image size is $320\times 320\times 3$.
      }
    \begin{tabular}{lccc}
        \toprule
        &RT(ms) &GFLOPs &F1 \\
        \midrule
        $T=0$&44.8 &96.1 &93.2 \\
        $T=1$&72.9 &186.8 &93.6 \\
        $T=2$& 93.2 &195.8 & 94.0 \\
        $T=3$& 114.3 &204.8 & 93.9 \\
        $T=4$&137.1 &213.7 &94.0 \\
        $T=5$&160.9 &222.7 &94.0 \\
        CNN~w.~similar model size &66.2 &214.5 &93.6 \\ 
        \midrule
        FCNCD&32.7&100.2&84.2 \\
        ADCDNet&41.3&217.4&93.5 \\
        CSCDNet&214.8&65.9&90.1 \\
        ARPPNet&32.3&44.6&86.8 \\
        IFN&35.0&128.5&83.6 \\
        BIT&13.7&53.8&89.0 \\
        STANet&66.2&20.1&82.9 \\
        ChangeFormer&100.2&316.9&88.3 \\
        STMSNet & 38.5&24.5&49.3\\
        SFBNet &50.7 & 141.2 & 44.9\\
        RCTNet &73.3 &32.4 & 83.7\\
         \bottomrule
    \end{tabular}
 
    \label{rebuttaltab:gflops}
\end{table}

\subsubsection{\textbf{Results of model complexity}}
In \tableref{rebuttaltab:gflops}, we report the running time, GFLOPs and F1 of IDETs with $T\in\{0,1,\ldots,5\}$ on VL-CMU-CD. We can find that the GFLOPS and running time become higher as $T$ increases. The F1 values of our method increase from $T=0$ to $T=2$, dropping when continue increasing $T$. $T=2$ is a good choice for the proposed method. To demonstrate the effectiveness of IDET, we replace IDET~(T=2) with a CNN model containing four convolution blocks with convolution layer, batch normalization, and ReLU, which has much higher GFLOPS than IDET. As shown in \tableref{rebuttaltab:gflops}, we find that the CNN-based method has a lower F1 value than IDET~(T=2).
Additionally, we also give the running time and GFLOPs of the compared methods in \tableref{rebuttaltab:gflops}. CSCDNet has the longest running time with 214.8 ms. BIT has the shortest running time with 13.7 ms. Due to stacking four complex Transformer blocks, ChangeFormer has the highest GFLOPs of 316.9 among all the CD methods. SFBNet and RCTNet achieve the running time by 50.7 and 73.3, respectively. IDET~($T=2$) has moderate GFLOPs and achieves the highest F1 value than other compared methods.

\subsubsection{\textbf{Overfitting validation}}
We conduct cross-validation experiments across 3 datasets (\ie, VL-CMU-CD, CDNet, PCD) and compare with ADCDNet, CSCDNet, IFN, BIT in \figref{rebuttalfig:cross}. `CMU2PCD' means that we train detectors on VL-CMU-CD dataset and sample the testing images from the PCD dataset to evaluate. We can find that IDET has a higher generalization ability than all baseline methods.

\section{Conclusion}
In this paper, we conduct a comprehensive study on a crucial factor that significantly impacts change detection accuracy: the quality of the feature difference. We construct a fundamental feature difference-based change detection module and thoroughly analyze how the quality of the feature difference affects the final detection results. To bridge this gap and enhance the feature difference, we propose a novel module called the iterative difference-enhanced transformers (IDET). Additionally, we introduce the multi-scale IDET method to achieve even better refinement and leverage it for the final change detection.
To validate the effectiveness of our proposed method, we perform extensive experiments and conduct ablation studies using six large-scale datasets that encompass various application scenarios, image resolutions, and change ratios. Our results demonstrate that IDET surpasses all baseline methods, highlighting the critical importance of feature difference quality. Moreover, the ability of IDET to improve other change detection methods further underscores the significance of feature difference quality in this field.

\section{Acknowledgments}

This research is supported by the National Research Foundation, Singapore and Infocomm Media Development Authority under its Trust Tech Funding Initiative, the National Research Foundation, Singapore, and DSO National Laboratories under the AI Singapore Programme (AISG Award No: AISG2-GC-2023-008), and Career Development Fund (CDF) of Agency for Science, Technology and Research (No.: C233312028). Any opinions, findings and conclusions or recommendations expressed in this material are those of the authors) and do not reflect the views of National Research Foundation, Singapore and Infocomm Media Development Authority. 
This work is also supported by the National Natural Science Foundation of China under Grant No. 62302415, Guangdong Basic and Applied Basic Research Foundation under Grant No. 2022A1515110692, 2024A1515012822.

\bibliographystyle{unsrt}
\bibliography{ref}

\begin{IEEEbiography}[{\includegraphics[width=1in,height=1.25in,clip,keepaspectratio]{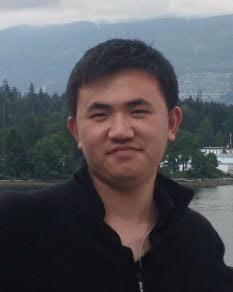}}]{Qing Guo}
	received his Ph.D. degree from the School of Computer Science and Technology, Tianjin University, China. He was a research fellow and the Wallenberg-NTU Presidential Postdoctoral Fellow at the Nanyang Technological University, Singapore, from Dec. 2019 to Sep. 2022. He is currently a senior research scientist and principal investigator at the Center for Frontier AI Research (CFAR), A*STAR in Singapore. He is also an adjunct assistant professor at the National University of Singapore (NUS). He serves as the Senior PC for AAAI and Area Chair for ICLR 2025. His research mainly focuses on computer vision, AI security, adversarial attacks, and robustness. He is a member of IEEE. 
\end{IEEEbiography}

\begin{IEEEbiography}[{\includegraphics[width=1in,height=1.25in,clip,keepaspectratio]{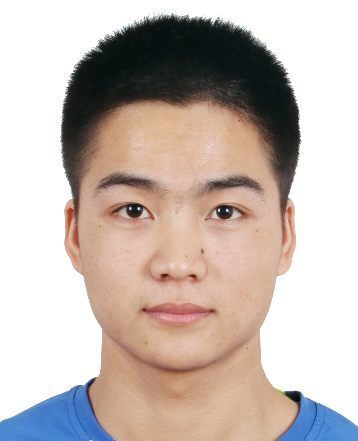}}]{Ruofei Wang}
    received  his  MS  degree from the School of Computer Science and Technology, Civil Aviation University of China in 2023. He is currently working toward a Ph.D. degree under the supervision of Dr. Renjie Wan in department of computer science, Hong Kong Baptist University. His research interests include change detection, neuromorphic vision, and AI security.
\end{IEEEbiography}

\begin{IEEEbiography}[{\includegraphics[width=1in,height=1.25in,clip,keepaspectratio]{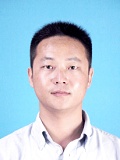}}]{Rui Huang}
        received the PhD from the Tianjin University under the supervised of Professor Wei Feng and Professor Jizhou Sun. Currently, he is a lecturer at the School of Computer Science and Technology in the Civil Aviation University of China. His research interests include computer vision, machine learning and visual surveillance.
\end{IEEEbiography}

\begin{IEEEbiography}[{\includegraphics[width=1in,height=1.25in,clip,keepaspectratio]{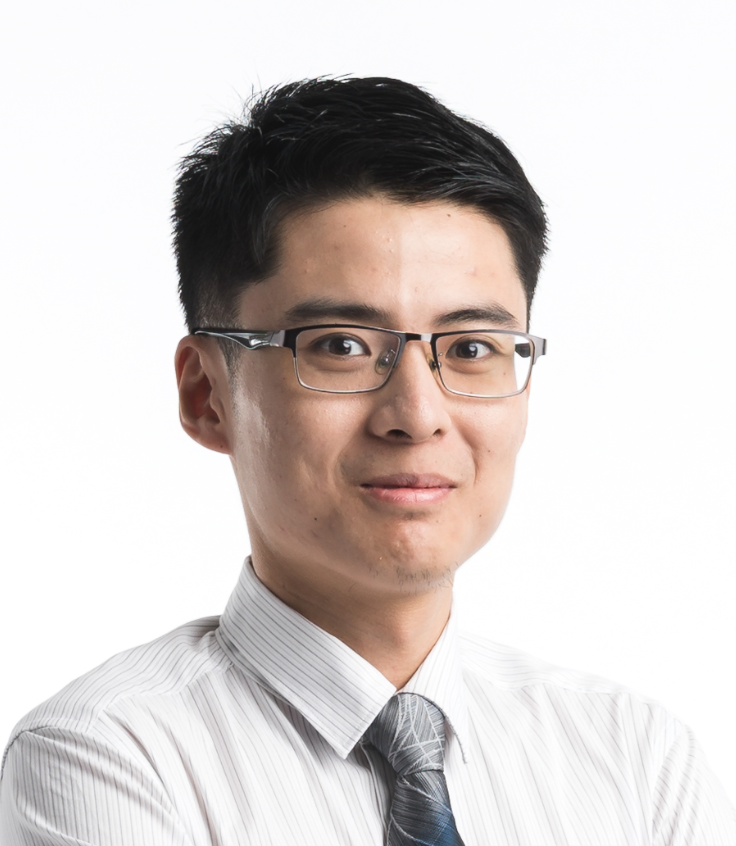}}]{Renjie Wan}
    received the B.Eng. degree from the University of Electronic Science and Technology of China in 2012 and the Ph.D. degree from Nanyang Technological University, Singapore, in 2019. He is currently an Assistant Professor with the Department of Computer Science, Hong Kong Baptist University, Hong Kong. He was a recipient of the Microsoft CRSF Award, the 2020 VCIP Best Paper Award, and the Wallenberg-NTU Presidential Postdoctoral Fellowship. He is the outstanding reviewer of the
2019 ICCV.
\end{IEEEbiography}

\begin{IEEEbiography}[{\includegraphics[width=1in,height=1.25in,clip,keepaspectratio]{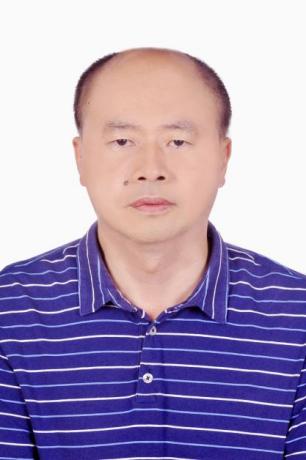}}]{Shuifa Sun}
        received the M.S. degree from Zhejiang University of Technology, China, in 2002, and the Ph.D. degree from Zhejiang University in 2005, respectively. He is now a professor of School of Information Science and Technology (SIST), Hangzhou Normal University, Hangzhou, China. His research interests include intelligent information processing, computer vision, digital forensics, and multimedia information processing.
\end{IEEEbiography}

\begin{IEEEbiography}[{\includegraphics[width=1in,height=1.25in,clip,keepaspectratio]{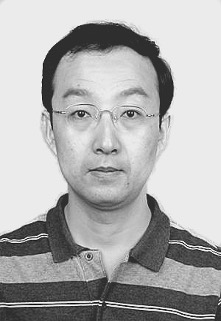}}]{Yuxiang Zhang}
        is currently an full professor in the School of Computer Science and Technology at Civil Aviation University of China. He received a Ph.D. degree from School of Electronic and Information Engineering, Beijing Jiaotong University, China, in 2011. His main research interests include keyphrases extraction, sentiment analysis on texts, and social network analysis.
\end{IEEEbiography}

\end{document}